# How do you #relax when you're #stressed? A content analysis and infodemiology study of stress-related tweets

Son Doan[1], PhD; Amanda Ritchart[2], CPhil; Nicholas Perry[3], MS; Juan D Chaparro[1], MD; Mike Conway[4], PhD

[1]Deparment of Biomedical Informatics, University of California, San Diego, La Jolla, CA, United States
[2]Linguistics Department, University of California, San Diego, La Jolla, CA, United States
[3]Department of Psychology, University of Utah, Salt Lake City, UT, United States
[4]Department of Biomedical Informatics, University of Utah, Salt Lake City, UT, United States

**Corresponding Author:**
Son Doan, PhD
Department of Biomedical Informatics
University of California, San Diego
9500 Gilman Dr, La Jolla, CA, 92093
United States
Email: sondoan@gmail.com

## Abstract

**Background:** Stress is a contributing factor to many major health problems in the United States, such as heart disease, depression and autoimmune diseases. Relaxation is often recommended in mental health treatment as a frontline strategy to reduce stress, thereby improving health conditions. Twitter is a micro-blog platform that allows users to post their own personal messages (tweets), including their expressions about feelings and actions related to stress and stress management (e.g., relaxing). While Twitter is increasingly used as a source of data for understanding mental health from a population perspective, the specific issue of stress – as manifested on Twitter – has not yet been the focus of any systematic study.

**Objective:** To understand how and in what way people express their feelings of stress and relaxation through Twitter messages. In addition, we investigate automated natural language processing (NLP) methods to (1) classify stress vs. non-stress and relaxation vs. non-relaxation tweets and (2) identify first-hand experience – i.e., who is the experiencer – in stress and relaxation tweets.

**Methods:** We first performed a qualitative content analysis of 1326 and 781 tweets containing the keywords "stress" and "relax", respectively. We then investigated the use of machine learning algorithms – in particular naïve Bayes and Support Vector Machines (SVMs) – to automatically classify tweets as stress vs. non-stress and relaxation vs. non-relaxation. Finally, we applied these classifiers to sample data sets drawn from four cities (Los Angeles, New York, San Diego, and San Francisco) obtained from Twitter's Streaming

Application Programming Interface (API), with the goal of evaluating the extent of any correlation between our automatic classification of tweets and results from public stress surveys.

**Results:** Content analysis showed that the most frequent topic of stress tweets was education, followed by work and social relationships. The most frequent topic of relaxation tweets was rest and vacation, followed by nature and water. When applied to the cities dataset, the proportion of stress tweets in New York and San Diego is substantially higher than in Los Angeles and San Francisco. In addition, we found that characteristic expressions of stress and relaxation vary for each city based on its geo-location.

**Conclusions:** This content analysis and infodemiology study revealed that Twitter, when used in conjunction with NLP techniques, is a useful data source for understanding stress and stress management strategies, and can potentially serve as a supplement to infrequently collected survey-based stress data.

## Introduction

Psychological stress has been linked to multiple health conditions, including depression [1], heart disease [2], autoimmune disease [3], and general all-cause mortality [4]. Stress has also been associated with worse health outcomes among those living with chronic illness [5], suggesting that stress may exacerbate pre-existing health conditions, as well as contribute to the development of new health problems. Stress not only contributes to physical and mental health problems, such as heart disease, depression, and autoimmune diseases [6], but also has negative impacts on family life and work, significantly impairing quality of life [7,8]. Accordingly, stress is an important concern for public health prevention initiatives [7,8].

Health surveys have demonstrated that stress negatively impacts a large proportion of the US population [9]. Underscoring the magnitude of the problem, a study conducted by Harvard School of Public Health found that 49% of the American public reported being stressed within the last year, and also found that that 60% of those who reported being in poor health also reported experiencing a substantial amount of stress within the last month [7]. Further, levels of stress appear to be unequally distributed throughout the population [10]. National surveys have documented that higher levels of stress are reported among those who were of lower income, less educated, and younger [11]. Theorists have suggested that geographic clustering of psychological characteristics may be driven by selective migration (i.e., in this case, people more vulnerable to stress seek out others like themselves), social influence (i.e., attitudes and beliefs that lead to greater stress cluster together geographically), or environmental influence (i.e., features of the physical environment, such as neighborhoods, increase stress among those who live close to one another) [12]. In short, large-scale studies have documented both the high prevalence of stress within the US, as well as geographic clustering of psychological distress, suggesting that tracking symptoms of stress should ideally occur at both the national and local levels.

Relaxation is considered a key component of frontline stress management techniques, such as cognitive-behavioral stress management [13]. General stress management can include adaptive coping (e.g., distraction), physical relaxation strategies (e.g., diaphragmatic breathing), cognitive reappraisal (e.g., reconsidering the stressor from a different perspective), and mindfulness (i.e., increasing awareness of the present moment). These stress management strategies are intended to reduce psychological and physiological arousal related to stress, promote healthier coping alternatives, and, in turn, reduce some of the negative health impacts of stress. Indeed, these strategies have been found to be effective for improving health outcomes among those living with chronic illness [14–16], as well as for improving general mental health and quality of life [17,18].

Understanding what the major causes of stress are and how people negatively or positively manage their stress (e.g., through stress management techniques such as cultivating relaxation) is important [7,19]. Population health surveys often use telephone interviews or questionnaires from samples of the population, e.g., CDC's Behavioral Risk Factor Surveillance System (BRFSS) [20]. These methods, although reliable, are conducted relatively infrequently due to cost, and may be less effective at reaching certain populations, such as those without a dedicated landline telephone. With the rapid growth of online social networks today, social media data can serve as a useful additional resource to understand aspects of stress that are difficult to assess in general surveys or clinical care. For example, social media provide a means to rapidly and dynamically address new and evolving research questions with a degree of flexibility not possible with surveys. Social media may also provide insights into populations that may be underrepresented in surveys (depending on the demographics of the particular social media platform used). Thus, social media can potentially serve as a beneficial supplement to detailed surveys when understanding public health concerns.

Twitter – one of the most popular social media platforms – is a micro-blog service that allows users to post their own personal messages (a 'tweet' with a 140-character limit). As of May 2016, it had 310 million active users with 1 billion unique visits monthly to sites with embedded tweets [21]. The utility of Twitter as a data source has been investigated in numerous applications such as election prediction [22], stock market prediction [23], oil price changes [22], and earthquake and disasters [24].

Twitter has also been used in public health for influenza tracking [25–27], studying breast cancer prevention [28], childhood obesity [29], issues related to general health [30], tobacco and e-cigarette use [31], dental pain [32,33], general pain [34], sexually transmitted diseases [35], and weight loss [36]. There has also been research regarding the general well-being of people in different geographical locations using Twitter messages [37], a correlation study of Twitter messages with depression [38], as well as with heart disease mortality [39]. However, no studies specifically focused on stress and stress management have been conducted until now.

In this paper, we investigate how and in what ways people express their own stress and relaxation through an in-depth content analysis of Twitter messages. In addition, we investigate automated methods to classify stress and relaxation tweets using machine learning techniques. Furthermore, we rank stress and relaxation levels based on the relative proportions of stress and relaxation related tweets (as identified by our NLP classifiers) in four U.S. cities: New York, Los Angeles, San Diego, and San Francisco. We then compare these results to public surveys reported by Forbes and CNN [40,41]. This study will provide another perspective on how people think about and cope with stress using easily-acquired, naturalistic Twitter data, complementing existing survey-based epidemiological methods.

## Methods

### Data Collection

#### Dataset 1

To begin our investigation of stress and relaxation (stress management) tweets, we first collected tweets with user-defined stress and relaxation topics using the Twitter REST Application Programming Interface (API) [42]. The user-defined topics included the hashtagged topics #stress and #relax, as well as variations of these words. The full search list used can be found in **Table 1**. Tweets were collected between July 9 and July 14, 2014. We supplemented this seed dataset with tweets from the random sample stream Twitter Streaming API [43] (1% sample rate) in order to have better representation of "everyday" tweets that did not necessarily contain stress and relaxation related hashtags, but that still contained the keywords "stress" or "relax." This dataset consists of 1326 stress-related and 781 relaxation-related tweets. We refer to this dataset as Dataset 1.

#### Dataset 2

We further investigated the characteristics of stress and stress management by geographical location (four cities) and compared the locations against each other using Dataset 2. This dataset – much larger than Dataset 1 – consisted of geo-tagged tweets obtained from the Twitter Streaming API [43] in one of four possible cities: Los Angeles,

**Table 1.** List of hashtags related to stress and relaxation to create Dataset 1.

| Stress-related hashtags | Relaxation-related hashtags |
|:---:|:---:|
| #stress | #relaxed |
| #stressed | #relaxin |
| #stressful | #relaxing |
| #stressin | #sorelaxin |
| #stressing #sostressful | #sorelaxing |

| |
|---|
| #sostressed |
| #stressinout |
| #stressingout |

New York, San Diego and San Francisco. These cities were chosen because they are densely populated and major metropolitan areas on the East and West Coasts of the United States. Tweets were collected between September 30, 2013 and February 10, 2014. The number of tweets for each city for this time period was 8.2 million for New York, 6.6 million for Los Angeles, 3 million for San Diego, and 4.4 million for San Francisco. Note that the most populous cities – i.e., New York and Los Angeles – generated the greatest number of tweets during the study period. We refer to this dataset as Dataset 2.

## Gold Standard and Manual Analysis of Tweets

Since our primary goal in this study is to understand how people express stress and relaxation through Twitter, we developed annotation guidelines for both stress and relaxation tweets based on reports from the American Psychological Association (APA) [7], Centers for Disease Control and Prevention (CDC) [8,44], and medical websites [6,45,46]. Following these guidelines, tweets were classified by both genre and theme. Genre reflects the format of the tweet (for example, personal experience), and theme reflects the domain of the actual content conveyed (including such categories as stress symptoms and stress topics).

Details for each genre and theme for stress and relaxation tweets are given below:
− Genre: We categorized tweets as being first-hand experience vs. other genre. First-hand experience was defined as a direct personal experience, or an experience directly related to the user writing the tweet. Other genres included second-hand experience, advertisements, news articles, etc. This genre classification was based on previous work on classifying health-related tweets [31]. After classifying a tweet as first-hand experience, we assigned its content into two themes: stress or relaxation.
− Stress themes: Content analysis focused on three main questions: (1) What kind of stress was being experienced? (2) What was the cause of the stress? and (3) What kind of actions, if any, were being taken regarding the stress? Based on these questions, we categorized the theme into three categories: stress symptoms, topics, and action(s) taken.
  • Symptoms: There were three classes of symptoms: (1) psychological and emotional, (2) physical, and (3) behavioral. These categories were based on guidelines for stress symptoms [47–49].
  • Topics: The general topic of a tweet, including (1) work, (2) education, (3) finances, (4) social relationships, (5) travel, (6) temporal, and (7) other. These topics were identified based on an analysis of data from Dataset 1.

- Action taken: This theme indicated the action that people reported taking when they were stressed. The action could be either negative or positive. An example of a negative action is: *I need a drink tonight. #sostressed.* An example of a positive action is: *I need a nap, and a hug. #stressingout #tired*.
- Non-specific: This theme was used for users that simply tweeted without any symptom, topic, or action. Examples include *#stressed!!!*, *Bad Night :,( #SoStressed*, etc.

— Relaxation themes: We categorized first-hand experience relaxation tweets by the topics (themes) given below.
- Topics: The action reported being taken by the user in order to relax, such as exercising or listening to music. A total of 11 topics were created based on data from Dataset 1: (1) physical, (2) water, (3) self care, (4) alcohol & drugs, (5) entertainment & hobbies, (6) food & drink, (7) nature, (8) rest & vacation, (9) social relationships, (10) other, and (11) non-specific.

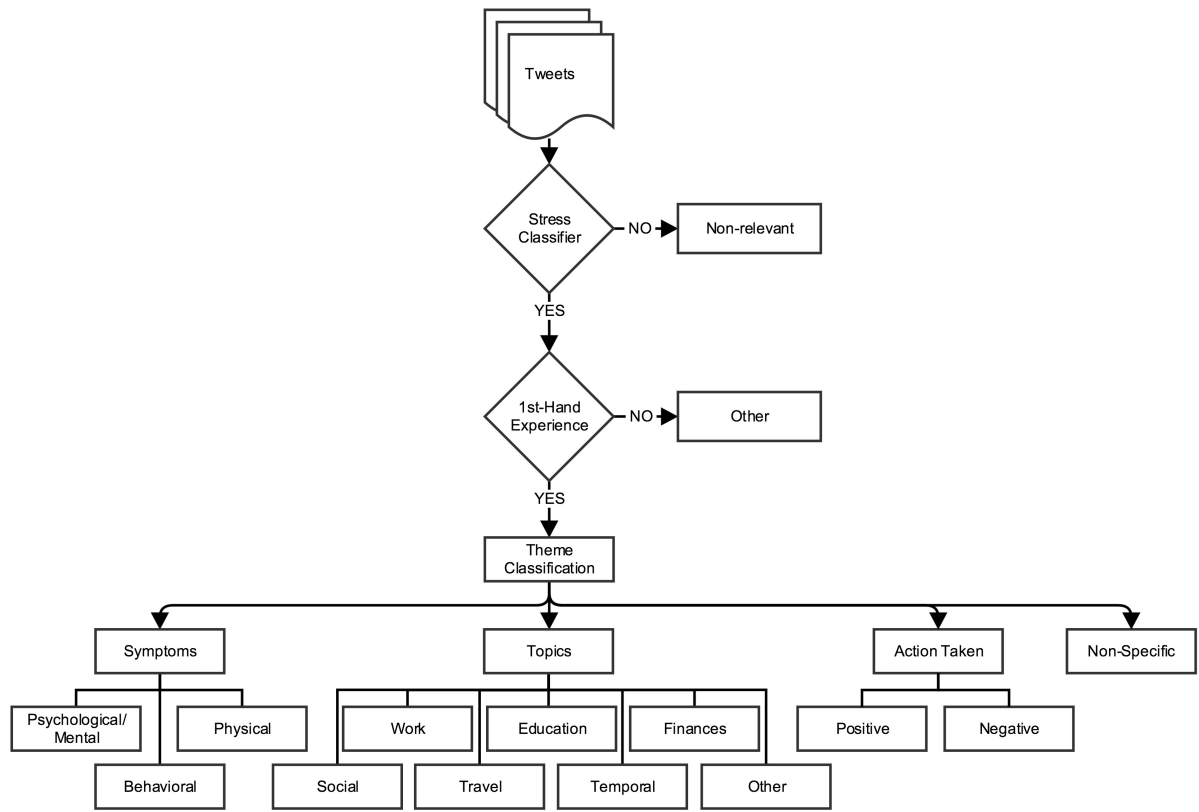

**Figure 1.** Schema used to classify stress tweets.

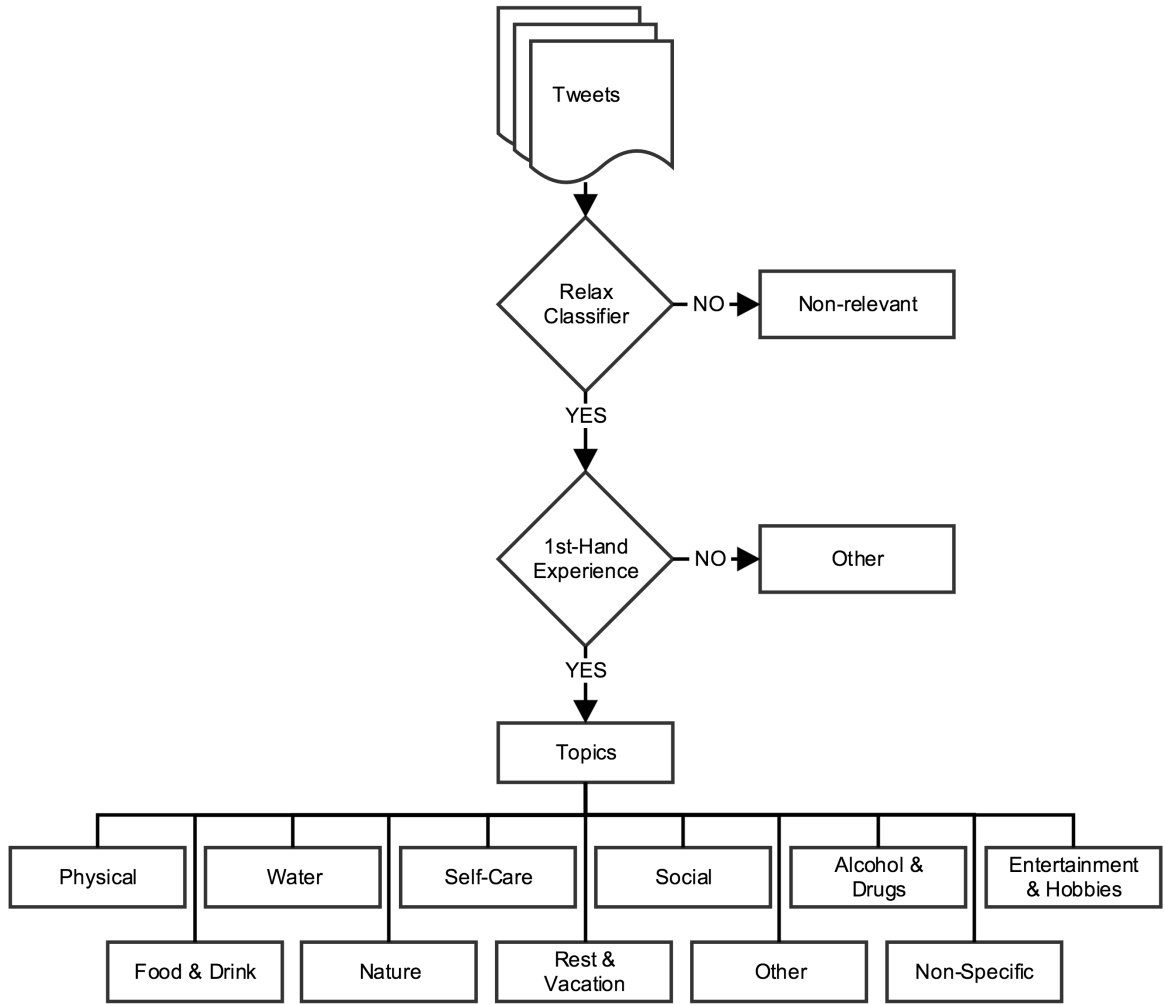

**Figure 2.** Schema used to classify relaxation tweets.

The schemas for stress and relaxation tweets are depicted in **Figure 1** and **Figure 2**. Definitions and examples of each category of first-hand experience tweets and its themes for stress and relaxation tweets are listed in **Appendix 1** and **Appendix 2**.

One author (AR) annotated stress and relaxation tweets from Dataset 1 and another (SD) annotated and verified the dataset to ensure that all tweets were annotated correctly. Any disagreements were resolved by meetings or exchanging emails. Dataset 1 contained a total of 664 stress and 662 non-stress tweets among the 1326 stress-related tweets, and a total of 391 relaxation and 390 non-relaxation tweets among the 781 relaxation-related tweets. For each stress or relaxation tweet, two authors (AR, SD) discussed and manually annotated tweets based on the guidelines as described above. After annotation, there were a total of 479 stress tweets and 335 relaxation tweets related to first-hand experience in Dataset 1. The details of Dataset 1 are depicted in **Figure 3**.

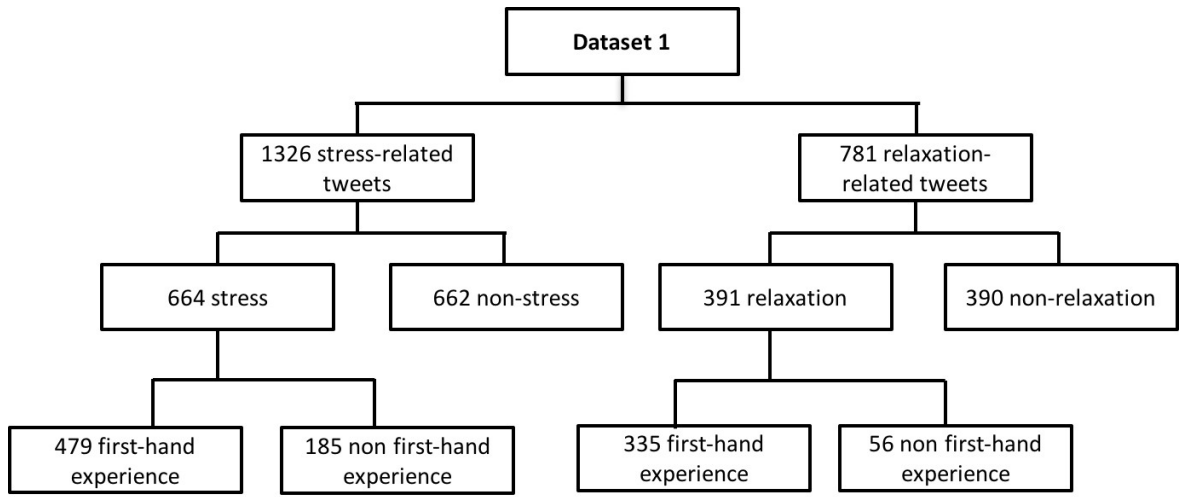

**Figure 3.** Description of Dataset 1.

Since the prevalence of some of the stress themes (e.g., *finances*, *work*) and relaxation themes (e.g., food *& drink*, *social*) in Dataset 1 was very low (i.e. too infrequent to train a machine learning classifier), we developed an automatic keyword-based theme classifier using a manually crafted lexicon of stress and relaxation keywords associated with each category. We first generated unigrams and bigrams from Dataset 1, and one author (AR) manually reviewed and selected the highest frequency unigram and bigram keywords. We then manually added corresponding synonyms into each theme in order to increase the coverage of the classifier. For example, the topic "education" in the stress schema contained unigrams "school", "college", "classes", and the bigram "high school" in Dataset 1. We manually added synonyms of those terms such as "exams" and "studying" as unigram keywords and "college life", "my tuition" and "on finals" into bigram keywords. The list was iteratively reviewed and confirmed by another author (SD). There was an average of 20 unigram and 20 bigram terms for each theme. Only unigram and bigram keywords were created since tweet messages are short in nature. Bigram keywords were necessary to include idiomatic expressions like "vicious cycle" and "hate feeling", and they also added more specificity such as "my heart" and "my sanity", which helped to increase the accuracy of the classifiers.

### Machine Learning Algorithms

Leveraging the annotated data derived from our content analysis of Dataset 1, we applied and evaluated machine learning algorithms for classification of stress vs. non-stress tweets and relaxation vs. non-relaxation tweets (on Dataset 1). In order to apply the classifier trained on Dataset 1 to the unseen, much larger Dataset 2 (cities dataset), we first filtered tweets by only keeping tweets that contained stress/relaxation-related hashtags in **Table 1** or keywords "stress"/"relax" for each city in Dataset 2. After this step, Dataset 2 contained only tweets with stress/relaxation-related keywords or hashtags. To calculate the proportion of stress/relaxation tweets at the city level, we utilized the stress/relaxation classifier trained on Dataset 1 to filter stress/relaxation tweets and then applied the classifier for first-hand experiencer to tweets from each city in Dataset 2. **Figure 4** shows a flow chart describing our machine learning design.

The work described in this paper focused on two machine learning-based classification tasks. First, tweets were classified into the appropriate stress and relaxation category (i.e., is it stress or relaxation related?). Second, first-hand experience tweets vs. non first-hand experience tweets were classified. We used two machine learning algorithms: naïve Bayes and Support Vector Machines (SVMs), which were implemented on Dataset 1 using 10-fold cross-validation. We used both the Naive Bayes and SVM algorithms, as both these algorithms have been used extensively for text classification tasks [50–52]. We used the *rainbow* package [51] for implementing both naïve Bayes and SVMs (linear kernel). We used "bag-of-words" as feature sets for both algorithms. The reason we used the "bag-of-word" representation is that this feature representation is considered as a baseline and the most common text representation in text classification in general [50–52]. To the best of our knowledge, this is the first study on classifying tweets on stress and relaxation tweets.

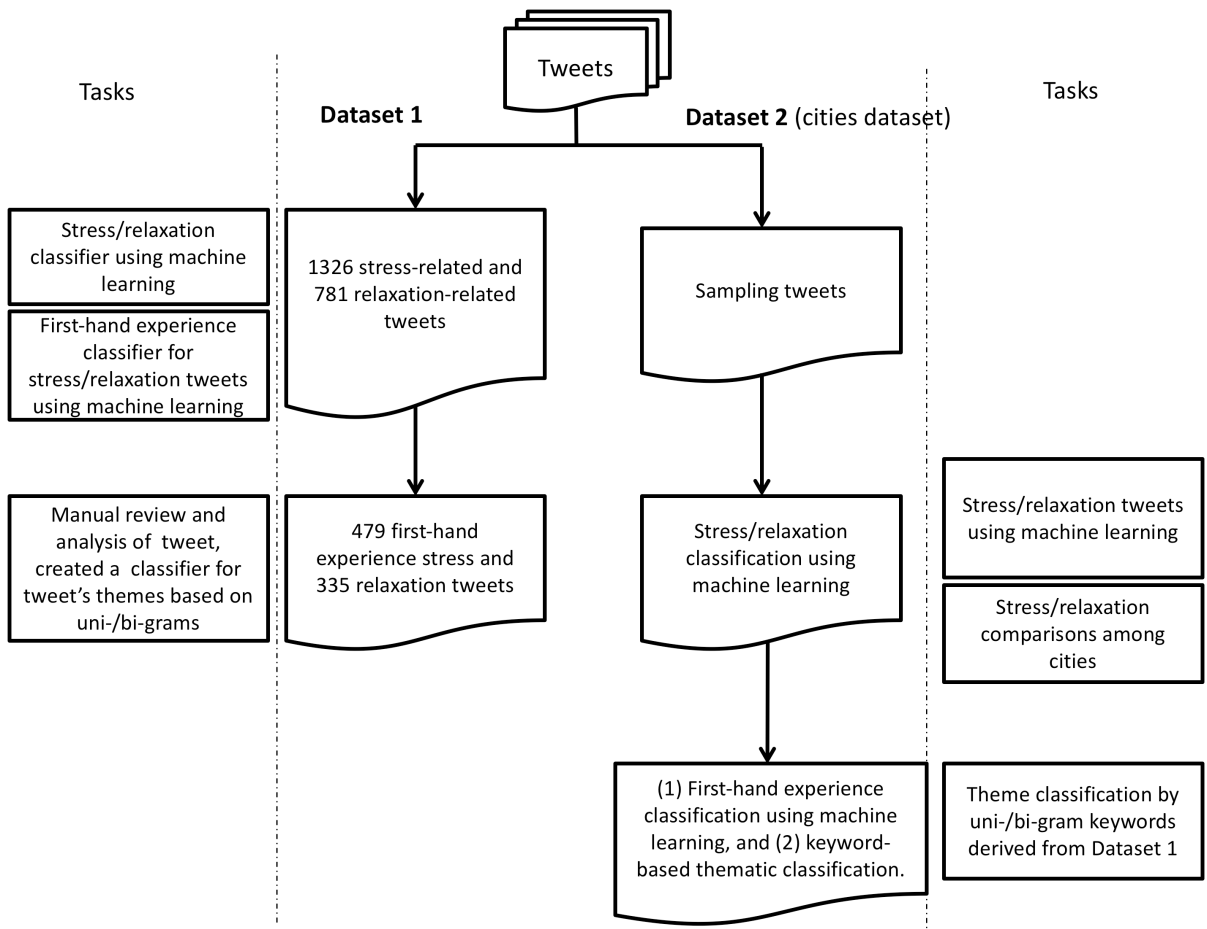

**Figure 4.** Datasets and tasks used for this study.

### Calculating Proportion of Stress and Relaxation Tweets at the City Level

We applied the two-step classification to each city in Dataset 2 to automatically identify stress and relaxation tweets. We calculated the proportion of stress/relaxation tweets to the total number of tweets in each city.

### Measurements and Statistical Analysis

For both stress/relaxation and first-hand experience classifications, we used accuracy, sensitivity, specificity, and Positive Predicted Values (PPV) as metrics [53–55] . They are defined as follows:

Sensitivity = TP/(TP + FN)
PPV = TP/(TP + FP)
Specificity = TN/(FP + TN)
Accuracy = (TP + TN)/(TP +TN + FP + FN)

Where TP is the number of tweets that are correctly classified as true, FP is the number of tweets that are incorrectly classified as true, FN is the number of tweets that are true but incorrectly classified as true, and TN is the number of tweets that are correctly classified as false.

In order to compare data among cities, we used Pearson's chi-squared test and reported significance if the P-value was less than 0.05 [56]. Statistical analyses were performed using the R package software, publicly available at https://www.r-project.org. Note that in order to preserve the anonymity of Twitter users, all example tweets reported in this paper are paraphrases of original tweets.

### Results

#### Content Analysis in Stress and Relaxation Tweets (Dataset 1)

**Figure 5a** shows the distribution of themes in first-hand experience stress tweets. This figure indicates that the highest frequency theme in stress tweets is topic, followed by symptoms (e.g., *Not sure what to do... #stressed #worried #lost*), non-specific (e.g., *#stressed!!!*), and action taken (e.g., *I need a drink #sostressed*). This suggests that Twitter users who post about stress usually post more about the cause or topic of their stress and less about actions and symptoms associated with stress.

Among the total number of stress-related tweets, we found that the most frequent topic was education (15%), followed by work (9%) and social relationships (8%). This is interesting because many of Twitter's users are young people who attend school [57,58]. It seems that education and issues related to education, e.g., exams and finals, are of the utmost concern for Twitter users. Examples of the education topic include: *Never doing a session B math course ever again #sostressful* or *my exam in less than a month?! #stressing*. The topic distributions of first-hand experience stress tweets are depicted in **Figure 5b**.

Relaxation-related tweets encompass a wider range of topics than stress-related tweets. The most frequent topic of relaxation tweets was rest and vacation (36%), followed by nature (22%) and water (20%). Topic distributions of first-hand experience of relaxation tweets are depicted in **Figure 6**.

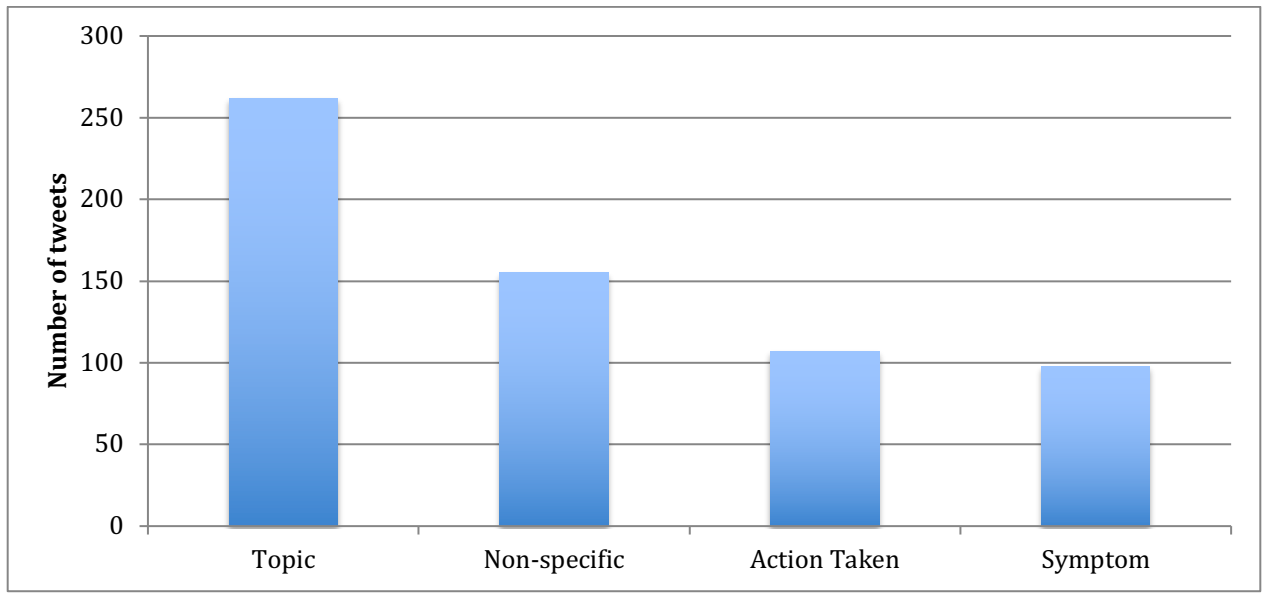

**Figure 5a.** Distributions by theme of first-hand experience stress tweets in Dataset 1.

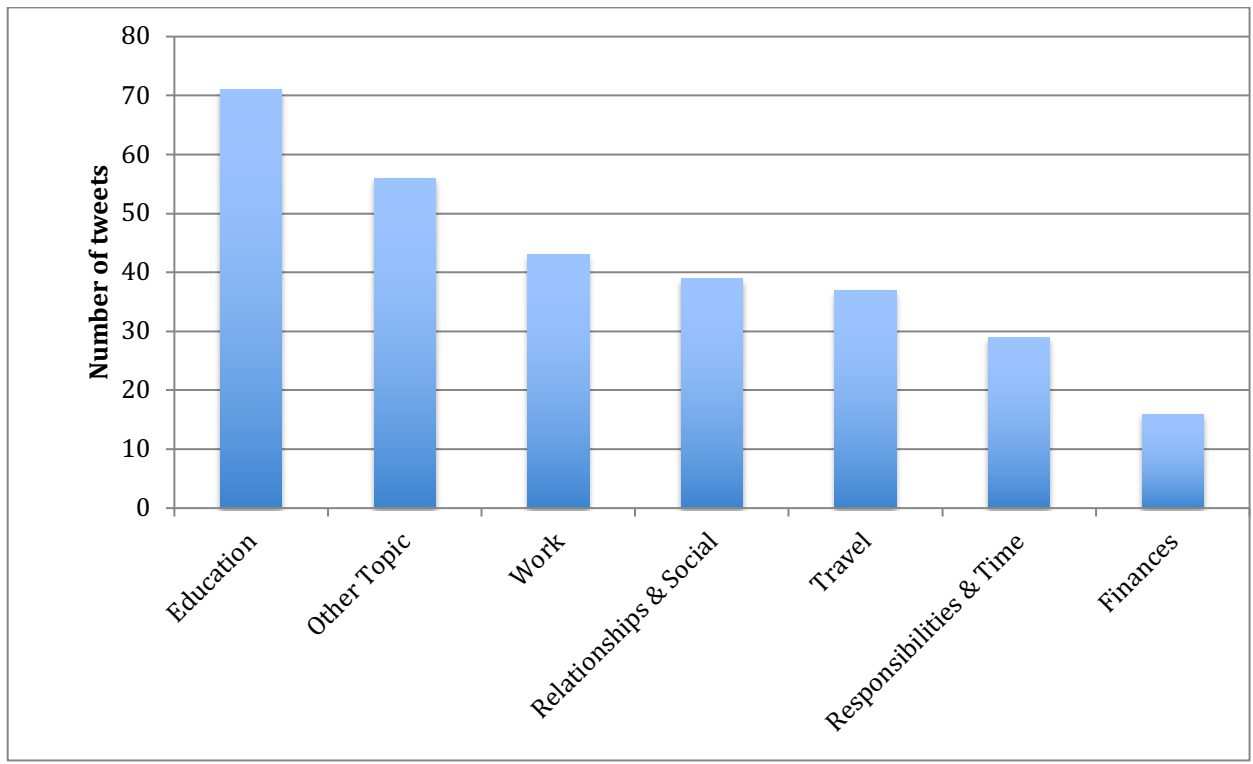

**Figure 5b.** Distributions by topics of first-hand experience stress tweets in Dataset 1.

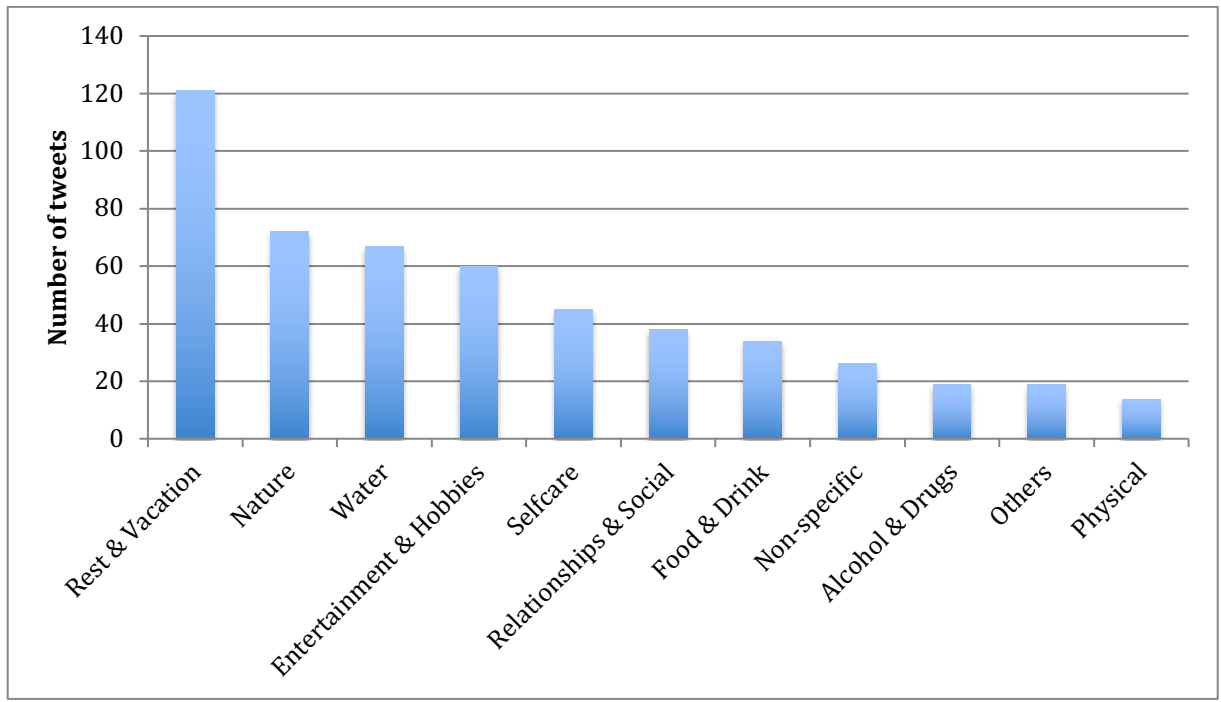

**Figure 6.** Distributions by topics of first-hand experience relaxation tweets in Dataset 1.

### Automatic Classification of Stress and Relaxation Tweets (Dataset 1)

Cross-validated classification results are shown in **Table 2**. Our results indicate that both algorithms achieved high accuracy (78.08-85.64%), sensitivity (90.26%-99.09%) and PPV (70.68%-89.32%), specificity is rather lower, especially with first-hand relaxation classification (naïve Bayes: 11.67%, SVM: 18.33%).

Of the two machine learning algorithms used, SVM (with linear kernel) performed better than naïve Bayes in classifying both stress vs. non-stress tweets (81.66% vs.78.64% accuracy, 92.73% vs. 91.97% sensitivity, 70.61% vs. 65.30% specificity, 76.07% vs.72.69% PPV). SVM is also better than naïve Bayes in classifying relaxation vs. non-relaxation tweets in accuracy (83.72% vs. 78.08%), specificity (77.18% vs. 60%) and PPV (79.86% vs. 70.68%) but slightly lower in sensitivity (90.26% vs. 96.15%).

**Table 2** also indicated that naïve Bayes had better accuracy and sensitivity in identifying first-hand experience stress and relaxation tweets: 87.58% vs. 85.61% (accuracy) and 95.53% vs. 90.64% (sensitivity) for stress; it achieved an accuracy of 85.64% vs. 83.85% and a sensitivity of 99.09% vs. 95.76% in comparison to SVM for relaxation tweets. In contrast, SVM performed better in specificity and PPV in classifying first-hand experience stress and relaxation tweets.

**Table 2.** Classification evaluation using 10-fold cross-validation on Dataset 1. We reported accuracy (Acc), Sensitivity (Sen), Specificity (Spec), and Positive predictive values (PPV) measures.

|  | Naïve Bayes Acc/Sen/Spec/PPV | SVM (linear kernel) Acc/Sen/Spec/PPV |
|---|---|---|
| Stress vs. Non-stress | 78.64/91.97/65.30/72.69 | 81.66/92.73/70.61/76.07 |
| Relaxation vs. Non-relaxation | 78.08/96.15/60.00/70.68 | 83.72/90.26/77.18/79.86 |
| First-hand vs. Non first-hand experience stress | 87.58/95.53/67.89/88.14 | 85.61/90.64/73.16/89.32 |
| First-hand vs. Non first-hand experience relaxation | 85.64/99.09/11.67/86.07 | 83.85/95.76/18.33/86.56 |

**Table 3.** Top 30 keywords ranked by information gain in stress and relaxation classification in Dataset 1.

| Stress vs. Non-stress | First-hand stress vs. Non-stress | First-hand relaxation vs. Non-relaxation | Relaxation vs. Non-relaxation |
|---|---|---|---|
| 0.03835 stressed | 0.06366 http | 0.08051 rt | 0.07980 rt |
| 0.03829 stress | 0.05561 rt | 0.07816 relaxing | 0.07615 relaxing |
| 0.03125 rt | 0.04562 stressed | 0.06342 relaxin | 0.05834 relaxin |
| 0.02143 mistress | 0.02878 stressing | 0.05032 sorelaxing | 0.04162 sorelaxing |
| 0.02137 stressful | 0.02246 stressful | 0.02835 relaxed | 0.03089 relaxed |
| 0.02030 stressing | 0.01956 mistress | 0.02064 work | 0.01808 time |
| 0.01265 http | 0.01766 stressingout | 0.01912 night | 0.01805 work |
| 0.01066 stressingout | 0.01089 sostressed | 0.01825 time | 0.01722 night |
| 0.01006 cashnewvideo | 0.01062 stressin | 0.01674 day | 0.01589 day |
| 0.00899 camerondallas | 0.00917 cashnewvideo | 0.01623 shower | 0.01444 cashnewvideo |
| 0.00736 burdenofstress | 0.00907 school | 0.01390 cashnewvideo | 0.01411 relax |
| 0.00633 tiger | 0.00881 ly | 0.01214 camerondallas | 0.01348 shower |
| 0.00612 stressin | 0.00830 stress | 0.01190 finally | 0.01261 camerondallas |
| 0.00605 sostressed | 0.00820 camerondallas | 0.01185 bath | 0.01127 relaxa |
| 0.00595 day | 0.00808 day | 0.01081 relax | 0.01079 video |
| 0.00580 nashgrier | 0.00778 love | 0.01076 listening | 0.01038 finally |
| 0.00580 distressed | 0.00690 sostressful | 0.01076 beach | 0.00984 bath |
| 0.00561 school | 0.00657 college | 0.01049 relaxa | 0.00974 home |
| 0.00558 anxiety | 0.00611 packing | 0.01038 video | 0.00894 vacation |
| 0.00555 life | 0.00588 life | 0.01012 home | 0.00894 listening |
| 0.00524 busy | 0.00577 twitter | 0.00967 vacation | 0.00894 beach |
| 0.00524 learn | 0.00577 tiger | 0.00967 pool | 0.00807 nashgrier |
| 0.00474 woods | 0.00534 hours | 0.00859 sitting | 0.00804 relaxar |
| 0.00421 bitch | 0.00529 big | 0.00859 enjoying | 0.00804 pool |
| 0.00419 hours | 0.00529 nashgrier | 0.00859 watching | 0.00804 enjoying |
| 0.00419 packing | 0.00529 distressed | 0.00859 rain | 0.00714 rain |
| 0.00418 twitter | 0.00481 hate | 0.00776 give | 0.00714 long |
| 0.00418 haha | 0.00457 long | 0.00776 nashgrier | 0.00714 sitting |
| 0.00416 college | 0.00457 weeks | 0.00751 long | 0.00714 watching |
| 0.00390 love | 0.00457 figure | 0.00698 bed | 0.00637 nice |

**Table 3** showed the terms that have highest information gain for stress/relaxation classification. Interestingly, we found that most terms characteristic of the stress class are related to the term "stress" such as "stressed" or "stressin". In contrast, most terms characteristic of the relaxation class are "vacation", "water", or "beach," which are related to the topics as categorized in our relaxation schema.

**Automatic Classification of Stress and Relaxation Tweets at the City Level (Dataset 2)**
Using a SVM algorithm trained on our annotated data (Dataset 1), we automatically classified the much larger Dataset 2 (cities dataset). We used a three step classification process. First, we filtered by keywords "stress"/"relax". Second, we applied the stress/relaxation classifier to this filtered data. Third, we used the first-hand classifier to identify first-hand stress/relaxation tweets. In both steps, we used SVM (linear kernel) trained on Dataset 1 as the classifier. The reason we used SVM because it had advantages in stress/relaxation classification in comparison to naïve Bayes in the Dataset 1. The number of tweets after each step is shown in **Table 4**.

To evaluate performance of stress/relaxation classification in Dataset 2, we randomly sampled two sets of 100 tweets, with each set consisting of 100 tweets containing either keyword "stress" (Set 1) or "relax" (Set 2) from a city in Dataset 2. We chose New York for evaluation since New York had the greatest number of tweets. Then 100 tweets from Set 1 were manually annotated (conducted by author SD) as stress/non-stress and first-hand stress/non first-hand experience stress class. Similarly, 100 tweets from Set 2 were also manually annotated as relaxation/non-relaxation and first-hand relaxation/non first-hand experience relaxation class.

**Table 5** showed results of classification on Set 1 and Set 2 using the SVM algorithm. It indicated fair accuracy (66%-92%) and high PPV (84.62-100%), however it has lower sensitivity in first-hand stress classification (44%) and specificity in relaxation classification (57.14%). The results of the SVM algorithm in Dataset 2 are different when compared to Dataset 1, perhaps due to different data distribution. The descriptions of manual annotation on 100 random tweets of Set 1 and Set 2 are shown in **Figure 7**.

**Figure 8** shows the proportion of stress/relaxation tweets out of all tweets by city in Dataset 2. The number of stress tweets is two times more than the number of relaxation tweets, indicating that Twitter users are more likely to tweet about stress than relaxation.

To evaluate theme classification by keyword matching, we randomly sampled 50 classified tweets for each theme from New York. Manually review showed that keyword classification achieved a PPV from 60% to 90% for relaxation tweets and 40% to 80% for stress tweets. Themes that have high PPV in relaxation tweets are *alcohol & drugs* (94%), *entertainment & hobbies* (94%) and *water* (92%), while ones have lower PPV include *nature* (60%) and *food & drink* (78%). For stress tweets, themes have high PPV include *finances* (84%), *education* (82%), and *behavioral* (82%), while *travel* (50%) and *temporal* (62%) have lower PPV. The numbers of classified first-hand stress/relaxation tweets by theme for each city are shown in **Appendix 3.**

First-hand classification results from Dataset 2 showed that cities manifest a uniform pattern of stress and relaxation tweets. We found that the singular first person pronoun "I" was consistently used the most across all cities when expressing stress, found in ~4% of all stress tweets, while in relaxation tweets "I" was used less often (ranked 7), at around 2.4%. Details of the 30 highest frequency keywords in first-hand experience stress and relaxation tweets for Los Angeles, New York, San Diego, and San Francisco are shown in **Appendix 4.**

We also found that linguistic expressions of negation such as "not," "but," "don't, " or quantifying words such as "much" are among the thirty unigrams most characteristic of stress-related tweets. In addition, users often use emotionally-laden swear words when expressing stress. It is important to note however that the affective polarity of certain swear words can be highly context dependent ("it's shit" vs. "it's the shit") [59]. Relaxation tweets, on the other hand, tend to contain words indicating relaxation and time such as "relax," "home," "time," "day," "now." We found that "home" is among the highest frequency terms in relaxation tweets, as is "weekend". Tag clouds of stress and relaxation tweets for each city are depicted in **Appendix 5.**

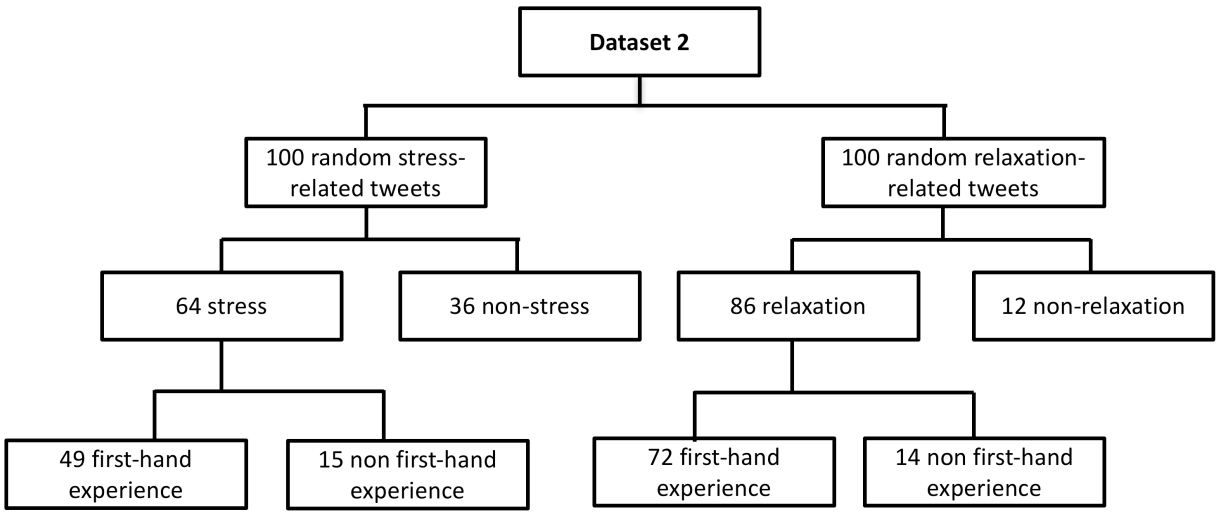

**Figure 7.** Description of manual annotation on 100 random tweets containing keywords "stress" and "relax" from Dataset 2.

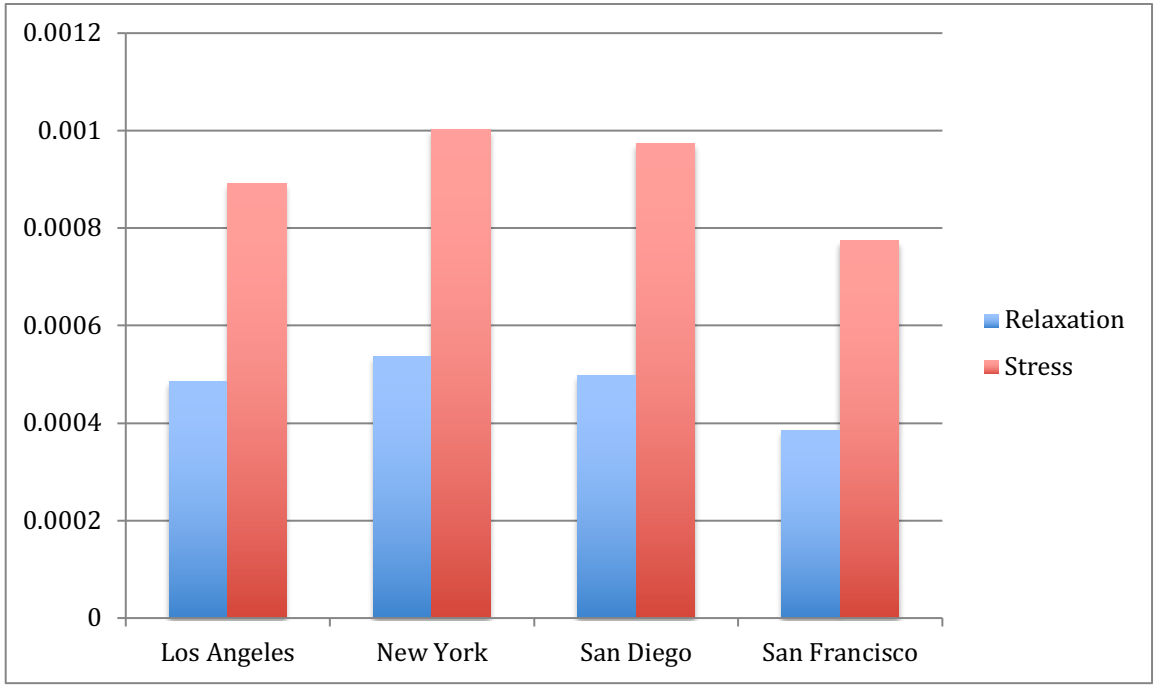

**Figure 8.** Proportion of relaxation/stress tweets by cities in Dataset 2.

**Table 4.** Stress ranking is based on 2011 Forbes [40] and 2014 CNN studies [41]. Statistical tests between cities showed there are differences between cities (P<0.0001), except San Diego and New York (Stress: P=0.18, Relaxation: P=0.02). P-values of relaxation and stress tweets between San Diego and Los Angeles are 0.41 and 0.000154, respectively. Ranks based on stress tweets are: New York=San Diego, Los Angeles, and San Francisco.

| Cities | Stress Rank 2011 (2014) | #tweets | #tweets contain "relax" | #tweets contain "stress" | relaxation tweets | stress tweets | relaxation tweets (first-hand) | stress tweets (first-hand) |
|---|---|---|---|---|---|---|---|---|
| Los Angeles | 1 (3) | 6,627,969 | 5,061 | 7,925 | 3,216 | 5,914 | 2,788 | 2,386 |
| New York | 2 (1) | 8,229,442 | 6,992 | 11,789 | 4,412 | 8,245 | 3,766 | 3,278 |
| San Diego | 5 (38) | 2,908,774 | 2,178 | 3,769 | 1,449 | 2,830 | 1,275 | 1,193 |
| San Francisco | 7 (39) | 4,372,966 | 2,554 | 4,558 | 1,682 | 3,384 | 1,471 | 1,389 |

**Table 5.** Classification evaluation using a random sample of 200 tweets (100 containing the keyword "stress"; 100 containing the keyword "relax") from New York in Dataset 2. We reported accuracy (Acc), Sensitivity (Sen), Specificity (Spec), and Positive predictive values (PPV) measures.

|  | SVM (linear kernel) Acc/Sen/Spec/PPV |
|---|---|
| Stress vs. Non-stress | 75.00/76.71/70.37/87.50 |
| Relaxation vs. Non-relaxation | 66.00/67.44/57.14/90.63 |
| First-hand vs. Non first-hand experience stress | 68.00/44.00/92.00/84.62 |
| First-hand vs. Non first-hand experience relaxation | 92.00/87.50/100.00/100.00 |

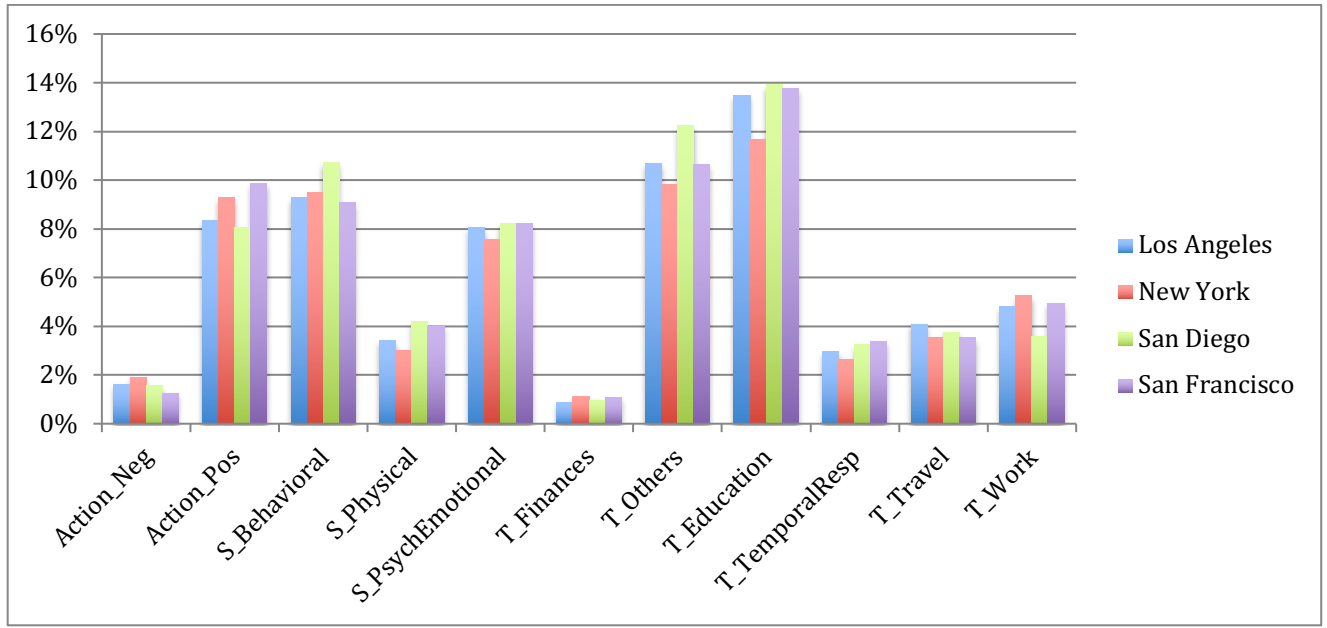

**Figure 9a.** Stress theme distribution by each of four cities in Dataset 2. Theme proportion of the first-hand experience stress tweets is labeled on the y-axis. There are no significant differences between cities (P>0.05).

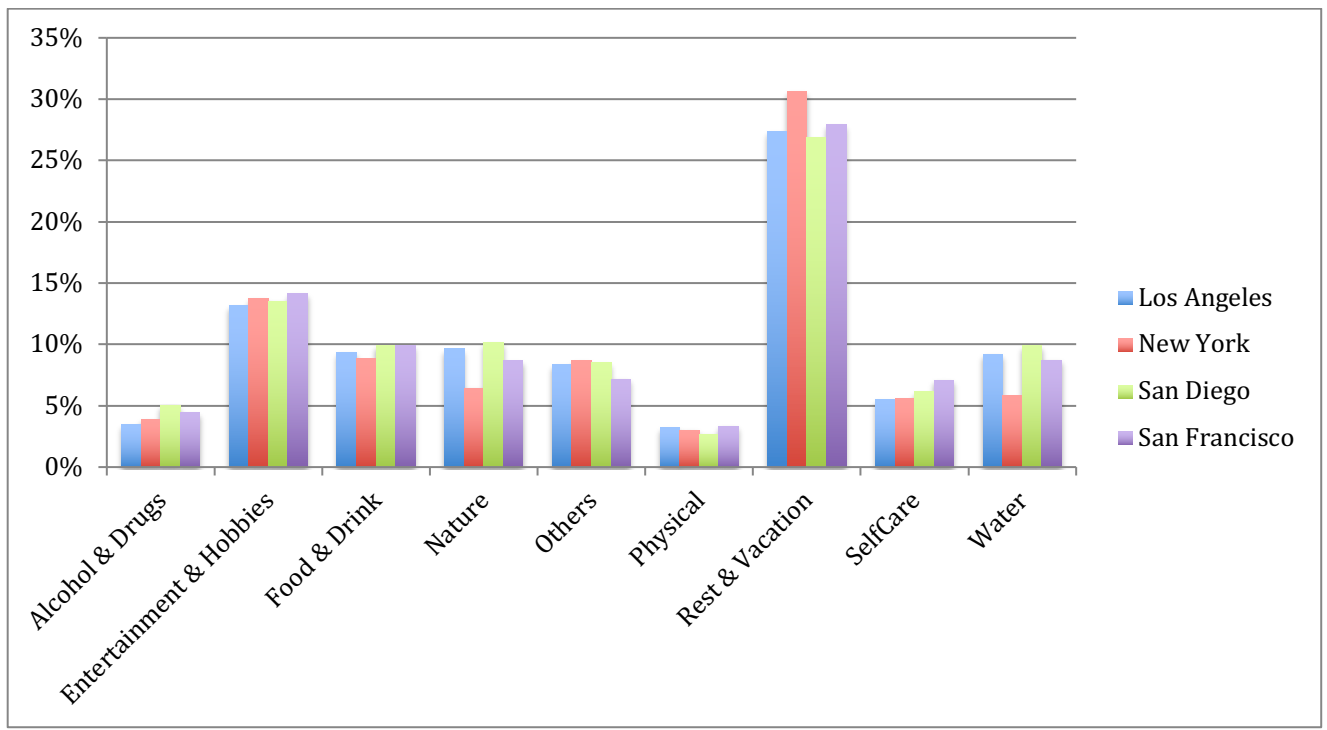

**Figure 9b.** Relaxation theme distribution by each of the four cities in Dataset 2. The y-axis is theme proportion of the first-hand experience relaxation tweets. There are significant differences between NY vs. other cities in the topics of Nature and Water.

### Theme Distributions of Tweets at the City Level (Dataset 2)

**Figure 9a** shows the theme distributions of stress tweets among cities. Education is the highest frequency topic (12-14%), followed by work (4-5%) and travel (4%). Interestingly, we found that tweets describing action taken and psychological & emotional symptoms also have relatively high frequencies (8-10%). This indicates that beside topic, people often post about their emotional state and reaction to stress.

The topic distributions of relaxation tweets are also consistent across cities. **Figure 9b** showed that rest & vacation is the highest frequency topic (27-31%), followed by entertainment & hobbies (13-14%), food & drink (9-10%), and nature (9-10%).

Though we do not find statistically significant differences in theme distributions among cities for stress tweets, there were significant differences between New York and other cities in the topics of nature and water in relaxation tweets. This may indicate the different activities taken for relaxation between the East Coast (New York) and West Coast (Los Angeles, San Diego, San Francisco). We found that high frequency terms for relaxation tweets in New York included "watching," while in San Diego "beach" was more common. This intuitively suggests that San Diegans more often relax by going to the beach, while New Yorkers relax by enjoying indoor (or spectator) entertainment ("watching", "listening").

### Correlations between Tweets Data Analysis and Public Surveys

Compared to two public surveys on the most stressful cities in the U.S. by Forbes [40] in 2011 and CNN [41] in 2014, the proportion of stress tweets found here are different. Both surveys ranked New York and Los Angeles among the most stressful cities in the country, while San Diego and San Francisco were categorized as less stressful. Our city rankings based on the proportion of first-hand experience stress tweets is New York followed by San Diego, Los Angeles, and San Francisco (**Table 6** and **Figure 8**). While we found no significant difference between New York and San Diego, we did find significant differences (P-value<0.0001) in pairwise comparisons between San Diego, Los Angeles, and San Francisco (**Table 6**).

Differences between results found in public stress surveys and our automatic classification of Twitter messages could be due to differences in methodology and population when collecting data. Public surveys collect data using telephones and paper-based reports, while Twitter messages are user-generated, naturalistic, and reflect personal thoughts. We suggest that Twitter could be used as a real-time, low-cost, and flexible supplement to public surveys when understanding and investigating stress and stress management techniques.

### Stress Relief by Relaxation in Tweets

The distribution of stress topics across cities shows an interesting finding: peoples' reactions to stress are more positive than negative. **Figure 9a** shows that for all cities, 8-10% of tweets report positive action taken in response to stress, while only 1-2% report negative action. This suggests that people may react to stress positively, or that people are more likely to publicly report positive rather than negative actions. Examples of positive reaction in stress tweets include rest (*Rest is best when you are stressed*) or exercising (*I'm so stressed, thank god I'm heading to yoga now*).

**Table 6.** P-values of pairwise comparisons of proportion of stress/relaxation tweets between the four studied cities.

| **Cities** | | Los Angeles | New York | San Francisco |
| --- | --- | --- | --- | --- |
| San Diego | Stress | 0.000154 | 0.18 | P<0.0001 |
|  | Relaxation | 0.405909 | 0.02 | P<0.0001 |
| San Francisco | Stress | P<0.0001 | P<0.0001 | NA |
|  | Relaxation | P<0.0001 | P<0.0001 | NA |
| New York | Stress | P<0.0001 | NA | P<0.0001 |
|  | Relaxation | P<0.0001 | NA | P<0.0001 |

Relaxation can be considered a stress management activity. **Figure 8** shows that the numbers of relaxation tweets are consistently proportional across all cities to those of stress tweets, indicating that Twitter users are consistently more inclined to post about stressful life events or experiences than relaxing experiences. Examples of stress relief from relaxation tweets include personal contact (*I don't need anything but a hug...*), exercising (*Went for a run, feel awesome, now time to relax*), shopping (*Last day in #SanDiego Just relaxing, shopping and say bye to friends*), or entertainment (*Relaxing watching a movie:-) :-)*). **Figure 6** and **9a** also indicate that rest & vacation is the highest frequency topic within relaxation tweets, followed by entertainment & hobbies, nature, and water. These topics can be considered common activities for stress relief.

## Discussion

### Principal Results
Our research addresses several aspects of the use of Twitter as a medium of expression of stress and relaxation by users. First, we created a schema for categorizing stress and relaxation-related tweets based on previously published psychological guidelines. By categorizing first-hand experience tweets into the primary themes of content topics, symptoms and actions taken, we gained further insight into the common patterns of expressions of stress.

Second, we analyzed in detail the contents of tweets based on our annotation scheme and found both similarities and differences in the prevalence and characteristics of stress and relaxation tweets across cities on the East and West Coasts of the United States. The most frequent topic of stress tweets in our datasets was education, which likely reflects the younger demographic of Twitter users [57,58], but work and travel were also common topics. It is notable that despite poverty rates, unemployment rates, and cost of living being significant factors in the methodology of CNN's and Forbes' stress ranking systems of most stressful cities, finances were not a major content topic of the stress tweets in any city. Although this result could be partially attributable to the need for either computer or mobile phone access in order to use Twitter and may cause under-representation in lower income groups, it may also indicate that certain topics, such as personal finances, still remain relatively taboo in social media settings. Regarding positive and negative actions regarding stress, positive actions far outnumbered more destructive behavior. The use of Twitter in itself to discuss feelings of stress and stress management can be seen as a constructive manner of dealing with stress by expressing these feelings and using the support of "followers" and friends. Social media platforms are increasingly being used as support networks in the management of chronic health conditions as varied as cancer, depression, and obesity. A recent systematic review by Patel et al. found that the impact of social media use on those experiencing chronic disease was positive in 48% of studies reviewed, neutral in 45%, and harmful in only 7% [60].

Third, our study indicates that words most associated with relaxation strategies (see **Table 3**) fall into three main groups: (1) bathing and personal care (e.g. "bath", "shower"), (2) vacationing ("vacation", "pool", "beach"), and (3) watching sports or TV ("videos", "sitting", "watching"), indicating that relaxation strategies involve purposefully taking time away from work-based activities and daily responsibilities. A further key theme that emerges from a qualitative analysis of the data is the idea of nature - in this case, particularly water (e.g. "pool", "beach", "rain") - as of key importance for relaxation. This result is consistent with recent research demonstrating the link between stress reduction and exposure to the natural environment (e.g. [61]).

Finally, we showed that machine learning algorithms could be employed to achieve good accuracy for the automatic classification of stress and relaxation tweets.

### Limitations

This study has several limitations. First, Dataset 2 was obtained from the Twitter API's 1% sample. Second, the annotation scheme we developed, although well suited for our purpose, could benefit from further refinement. For example, we found that many tweets were categorized as topic "other". Third, it is likely that classification results could be improved given the availability of additional training data, in particular for first-hand experience classification of stress and relaxation tweets. Furthermore, utilizing additional feature sets – e.g. ngrams, emotions, negations – could help improve accuracy. Fourth, Twitter reports of stress and relaxation may be influenced by self-presentation issues (e.g. stress related to excessive workload can be used as a status indicator in some contexts). Finally, as with all social media-based research, the population studied is unlikely to be a representative sample of the general population.

## Conclusions

In summary, this research shows that Twitter can be a useful tool for the analysis of stress and relaxation levels in the community, and has the potential to provide a valuable supplement to social and psychological studies of stress and stress management.

## Acknowledgements

We would like to thank Mr. Gregory Stoddard, MPH, MBA at the University of Utah's Division of Epidemiology for his valuable comments on an earlier version of this manuscript.

## Conflicts of Interest

None declared.

## Abbreviations

NLP: Natural Language Processing
SVM: Support Vector Machines
API: Application Programming Interface
APA: American Psychological Association
CDC: Centers for Disease Control and Prevention

# APPENDIX

**Appendix 1.** Examples of first-hand experience stress tweets with its themes. Sub-categories of themes are separated by a colon.
Note that all tweets have been paraphrased to preserve user anonymity.

**Stress: Symptom: Psychological & Emotional**

**Examples:**

| |
|---|
| No idea what to do...,, #stressed #worried #lost #frustrated |
| #Broken, #Dejected, #Stressed, #Depressed. |
| How do I deal with all this. #StressinOut |
| Fell asleep. Had nightmare. woke up stressed, and somehow my cat bit me. Can things go right for me just once? |

**Stress: Symptom: Behavioral**

**Examples:**

| |
|---|
| I need a drink tonight. #sostressed |
| I'm soooo stressed...I think I might go on an chocolate binge. |
| im so stressed. started smoking again. hadnt done it in a very long time |
| Haven't been able to sleep past 9 AM for the past 2 weeks and getting told to do things is just too stressful.. |

**Stress: Symptom: Physical**

**Examples:**

| |
|---|
| Chest hurts all nighttt #stressing |
| You get sick and are bedridden for only a few hours but because of the massive workload, you pretty much need to be dead. #stress |
| Feeling sick and soo stressed from school. Waiting for the weekend to lift my spirits. |
| Right now, Im shaking and my heart is beating so fast I can't even think! I am broken! Only God can heal me! #stressing |

**Stress: Symptom: Work**

**Examples:**

| |
|---|
| This job will kill me! #stress #nosleep |
| I'm annoyed with my job.. no vacation.. instead i have extra hours and a store without a manager #stressingOut |
| This is the first time I've cried for a long time, and all because of work. #sostressed |
| I'm not enjoying work much, to much stress! |

**Stress: Topic: Education**

**Examples:**

| |
|---|
| I just wanna finish my homework and go to sleep. This week is going to be horrible #StressingOut |
| College is #Stressful |
| Not doing a math course online again #sostressful #whatsgoingon |
| How can my exam be in less than a month?! #stressing |

**Stress: Topic: Finances**

| **Examples:** |
|---|
| I cannot wait till I get a decent pay check again #stressing |
| I don't know how I'm gonna afford everything  #stressingout |
| Why do you need to pay for class? #stressingout |
| I'm really tired of being broke!!!!! #Stressed |

**Stress: Topic: Relationships & Social**

| **Examples:** |
|---|
| Ain't nobody here for me but my kids. #stressingout |
| My husband going to trial tommorow #Stressingout |
| Omg my parents are being horrible.  i hate them. they put a lot of stress on me.  I get blamed for everything. |
| Freaking out when we don't know where mom is.. #stressingout |

**Stress: Topic: Responsibilities & Time**

| **Examples:** |
|---|
| Known for waiting until the last minute #SoStressful |
| So many things to do and not enough time to do it in... #StressingOut |
| Coaches said they want a decision Monday...just stressed now moe |
| So much to do! thought summers were for chilin #stressingout #needtorelax #ahhhh #cry |

**Stress: Topic: Travel**

| **Examples:** |
|---|
| Don't like last minute packing. #sostressful |
| is stressing, I have court today, but not enough gas, =( |
| why do I always feel so lost down here?? Stressing me out |
| I don't have enough gas – running out... #stressingout |

**Stress: Topic: Other**

| **Examples:** |
|---|
| When I was a kid, always wanted to be older #sostressful |
| cleaned my apartment. I've been messy lately I hope things get better. #stressed |
| Played this week for the 1st time. #stressin |
| prom is stressing me...and its weeks away! |

**Stress: Action: Negative**

| **Examples:** |
|---|
| I need some nicotine right now... #stressful |
| I hope my herbal cigarrettes arrive tomorrow, I'm stressed. |
| I need a drink tonight. #sostressed |
| Court in the morning – trying to drink away the stress |

**Stress: Action: Positive**

| **Examples:** |
|---|
| I need food, a sleep, and a hug. #stressingout #tired #hungry |
| Stressful day. #bath #relax |
| I need to go for a run, getting stressed by this class |
| smooth jazz is going to get me through the day. #sostressed |

**Stress: None-specific**

| **Examples:** |
|---|
| #stressed!!! |
| #Stressed about everything!! |
| Bad Night  #SoStressed... |

**Appendix 2.** Examples of first-hand experience relaxation tweets with its themes. Sub-categories of themes are separated by a colon.

**Relax: Physical**

| **Examples:** |
|---|
| off to spin class to get a good workout.  just want to relax:D |
| Yoga is my new thing #SoRelaxing |
| My wake up routine is doing 50 crunches, and going for a run. #Relaxing |
| Awesome workout.  I see a relaxing night ahead! |

**Relax: Water**

| **Examples:** |
|---|
| A bubble bath is wonderful after a bad day! #relaxed |
| Mom's pool for down time and tanning! #relaxin #poolside |
| Thunderstrom #SoRelaxing when you're inside! |
| Finally some #rain - weather cooled down #relaxed #cool |

**Relax: Self-Care**

| **Examples:** |
|---|
| Take a relaxing night in the hot tub soon. |
| What an amazing day! Going to have a massage.  incredible weather! #luckygirl #sorelaxing http://t.co/EhhTB2Sdeo |
| Just took a relaxing shower |
| Pedicures! #relaxing #happydays |

**Relax: Social**

| **Examples:** |
|---|
| Day off . #relaxing |
| Drinks on a #Monday Lunch coming up #relaxin |
| A relaxing, stress-free night. can't wait! |
| just got home from work busy day,glad its over #relax |

**Relax: Alcohol & Drugs**

| **Examples:** |
|---|
| morphine is great! #relaxin |
| Enjoying a beer, relaxing.. I deserve it! |
| Off work! I just wanna relax... |
| All you gotta do is put a #drinkinmyhand and I'm a happy person #relax |

**Relax: Entertainment & Hobbies**

**Examples:**

| |
|---|
| I'm gonna do laundry, go in the shower, and relax |
| #Relaxin doing what they do best |
| Smoking hookah and watching tV. #SoRelaxing |
| Best way to enjoy my day off. #Relaxing |

**Relax: Food & Drink**

| **Examples:** |
|---|
| Yesterday: Had a nice #bbq with #friends. We #relaxed, and had #fun. |
| drinking coffee and generally just relaxing until class begins |
| raisin bagel, painted nails, city and colour playing and the moon is right outside my window #relaxing |
| home after a long day, going to have a cup of tea and relax, good night all |

**Relax: Nature**

| **Examples:** |
|---|
| Day off at the beach #relaxin |
| Sitting here watching the rain #relaxed |
| I love the rain, helps me relax… I'm so tired |
| #moodchanger #relaxin #coolin |

**Relax: Rest & Vacation**

| **Examples:** |
|---|
| relaxing and looking at the sunset. Thank goodness for life and being able to breath. |
| #beach #summer #vacation #sunnyday #smile #loveit #relaxed |
| It's one of those days. #vacation #relaxed |
| The conference was amazing! Taking a day off to relax |

**Relax: Other**

| **Examples:** |
|---|
| Been sitting in my bed for an hour #sorelaxing |
| sometimes I like to just close my eyes...and relax |
| Bored #selfie #relaxin #lifestyle  ... |

**Relax: None-specific**

| **Examples:** |
|---|
| Relaxing |
| #relaxin |
| now I wait and relax |
| just relaxin now too |

**Appendix 3.**

**Appendix 3a.** Number of classified first-hand stress tweets by theme in each city, the first-hand stress tweets are classified using the SVM classifier.

| City | Action_Neg | Action_Pos | S_Behavioral | S_Physical | S_PsychEmotional |
|---|---|---|---|---|---|
| Los Angeles | 39 | 199 | 222 | 82 | 192 |
| New York | 62 | 304 | 311 | 100 | 248 |
| San Diego | 19 | 96 | 128 | 50 | 98 |
| San Francisco | 18 | 137 | 126 | 56 | 114 |

| City | T_Finances | T_Others | T_School | T_Temporal_Resp | T_Travel | T_Work |
|---|---|---|---|---|---|---|
| Los Angeles | 22 | 255 | 322 | 71 | 98 | 115 |
| New York | 37 | 322 | 383 | 88 | 117 | 172 |
| San Diego | 12 | 146 | 166 | 39 | 45 | 43 |
| San Francisco | 15 | 148 | 191 | 47 | 49 | 69 |

**Appendix 3b**. Number of classified first-hand relaxation tweets in each city, the first-hand relaxation tweets are classified using the SVM classifier.

| City | AlcoholDrugs | Ent_Hobbies | FoodDrink | Nature |
|---|---|---|---|---|
| Los Angeles | 98 | 368 | 262 | 269 |
| New York | 149 | 518 | 332 | 244 |
| San Diego | 65 | 173 | 127 | 130 |
| San Francisco | 66 | 209 | 146 | 128 |

| City | Others | Physical | Rest_Vaca | SelfCare | Water |
|---|---|---|---|---|---|
| Los Angeles | 234 | 91 | 764 | 155 | 258 |
| New York | 328 | 116 | 1153 | 212 | 219 |
| San Diego | 109 | 35 | 343 | 79 | 126 |
| San Francisco | 106 | 50 | 411 | 105 | 129 |

# Appendix 4

**Appendix 4a.** Top 30 highest frequency keywords in first-hand experience stress tweets for Los Angeles, New York, San Diego, and San Francisco.

| | | | | | | | | | | | |
|---|---|---|---|---|---|---|---|---|---|---|---|
| | | | | | | Stress category | | | | | |
| Los Angeles | | | New York | | | San Diego | | | San Francisco | | |
| i | 2692 | 4% | i | 3604 | 4.10% | i | 1216 | 3.90% | i | 1626 | 4.10% |
| stress | 2191 | 3.30% | stress | 2873 | 3.20% | stress | 968 | 3.10% | stress | 1198 | 3% |
| to | 1739 | 2.60% | to | 2332 | 2.60% | to | 840 | 2.70% | to | 1069 | 2.70% |
| the | 1473 | 2.20% | the | 2032 | 2.30% | the | 696 | 2.20% | the | 834 | 2.10% |
| out | 1330 | 2% | stressed | 2020 | 2.30% | stressed | 662 | 2.10% | stressed | 804 | 2% |
| stressed | 1330 | 2% | a | 1687 | 1.90% | out | 623 | 2% | out | 755 | 1.90% |
| my | 1228 | 1.80% | my | 1621 | 1.80% | is | 611 | 2% | my | 746 | 1.90% |
| and | 1185 | 1.80% | is | 1576 | 1.80% | my | 595 | 1.90% | and | 741 | 1.90% |
| is | 1144 | 1.70% | out | 1447 | 1.60% | so | 581 | 1.90% | is | 678 | 1.70% |
| a | 1136 | 1.70% | and | 1444 | 1.60% | and | 556 | 1.80% | a | 657 | 1.70% |
| so | 1072 | 1.60% | so | 1381 | 1.60% | me | 505 | 1.60% | so | 632 | 1.60% |
| me | 1031 | 1.50% | me | 1243 | 1.40% | a | 498 | 1.60% | me | 607 | 1.50% |
| i'm | 904 | 1.40% | this | 1129 | 1.30% | i'm | 459 | 1.50% | i'm | 541 | 1.40% |
| stressful | 870 | 1.30% | stressful | 1125 | 1.30% | stressful | 435 | 1.40% | stressful | 497 | 1.30% |
| this | 757 | 1.10% | i'm | 1095 | 1.20% | this | 384 | 1.20% | this | 477 | 1.20% |
| stressing | 655 | 1% | of | 833 | 0.90% | of | 319 | 1% | of | 396 | 1% |
| of | 640 | 1% | stressing | 811 | 0.90% | stressing | 313 | 1% | stressing | 385 | 1% |
| it | 585 | 0.90% | for | 752 | 0.80% | for | 284 | 0.90% | it | 357 | 0.90% |
| for | 543 | 0.80% | it | 749 | 0.80% | it | 263 | 0.80% | for | 325 | 0.80% |
| be | 499 | 0.80% | be | 738 | 0.80% | be | 244 | 0.80% | just | 299 | 0.80% |
| just | 490 | 0.70% | that | 661 | 0.70% | just | 239 | 0.80% | be | 295 | 0.70% |
| that | 459 | 0.70% | in | 638 | 0.70% | about | 224 | 0.70% | about | 280 | 0.70% |
| you | 452 | 0.70% | just | 630 | 0.70% | that | 216 | 0.70% | in | 272 | 0.70% |
| about | 444 | 0.70% | you | 551 | 0.60% | much | 206 | 0.70% | that | 267 | 0.70% |
| in | 440 | 0.70% | about | 535 | 0.60% | have | 196 | 0.60% | you | 249 | 0.60% |
| have | 374 | 0.60% | have | 497 | 0.60% | in | 192 | 0.60% | have | 220 | 0.60% |
| much | 374 | 0.60% | not | 494 | 0.60% | you | 190 | 0.60% | not | 214 | 0.50% |
| all | 362 | 0.50% | all | 490 | 0.60% | all | 185 | 0.60% | all | 213 | 0.50% |
| not | 350 | 0.50% | much | 485 | 0.50% | not | 171 | 0.50% | much | 212 | 0.50% |
| over | 340 | 0.50% | don't | 470 | 0.50% | but | 161 | 0.50% | but | 205 | 0.50% |

**Appendix 4b.** Top 30 highest frequency keywords in first-hand experience relaxation tweets for Los Angeles, New York, San Diego, and San Francisco.

| Relaxation category | | | | | | | | | | | |
|---|---|---|---|---|---|---|---|---|---|---|---|
| Los Angeles | | | New York | | | San Diego | | | San Francisco | | |
| relax | 1571 | 3.80% | relax | 2296 | 4.10% | relax | 706 | 3.80% | relax | 811 | 3.60% |
| to | 1397 | 3.40% | to | 1794 | 3.20% | to | 620 | 3.30% | to | 768 | 3.40% |
| and | 1266 | 3.10% | and | 1692 | 3% | and | 595 | 3.20% | and | 758 | 3.40% |
| a | 1152 | 2.80% | relaxing | 1463 | 2.60% | relaxing | 524 | 2.80% | a | 585 | 2.60% |
| relaxing | 1138 | 2.80% | the | 1346 | 2.40% | the | 489 | 2.60% | relaxing | 585 | 2.60% |
| the | 1057 | 2.60% | a | 1335 | 2.40% | a | 466 | 2.50% | i | 547 | 2.40% |
| i | 957 | 2.30% | i | 1317 | 2.40% | i | 405 | 2.20% | the | 526 | 2.40% |
| t | 653 | 1.60% | my | 817 | 1.50% | t | 371 | 2% | my | 318 | 1.40% |
| co | 650 | 1.60% | just | 714 | 1.30% | co | 366 | 2% | just | 287 | 1.30% |
| http | 644 | 1.60% | t | 685 | 1.20% | http | 361 | 1.90% | t | 281 | 1.30% |
| just | 565 | 1.40% | co | 680 | 1.20% | my | 248 | 1.30% | co | 277 | 1.20% |
| my | 537 | 1.30% | http | 672 | 1.20% | just | 247 | 1.30% | http | 275 | 1.20% |
| so | 418 | 1% | so | 538 | 1% | so | 185 | 1% | in | 228 | 1% |
| time | 412 | 1% | with | 537 | 1% | in | 184 | 1% | so | 214 | 1% |
| in | 411 | 1% | in | 535 | 1% | time | 183 | 1% | time | 214 | 1% |
| of | 389 | 0.90% | is | 523 | 0.90% | is | 171 | 0.90% | is | 209 | 0.90% |
| with | 377 | 0.90% | of | 505 | 0.90% | at | 161 | 0.90% | for | 200 | 0.90% |
| is | 356 | 0.90% | time | 467 | 0.80% | day | 157 | 0.80% | of | 198 | 0.90% |
| day | 341 | 0.80% | day | 453 | 0.80% | of | 156 | 0.80% | day | 190 | 0.90% |
| for | 329 | 0.80% | for | 432 | 0.80% | for | 151 | 0.80% | with | 182 | 0.80% |
| on | 308 | 0.80% | on | 427 | 0.80% | with | 146 | 0.80% | on | 166 | 0.70% |
| at | 304 | 0.70% | home | 424 | 0.80% | on | 142 | 0.80% | at | 159 | 0.70% |
| it | 250 | 0.60% | at | 360 | 0.60% | it | 115 | 0.60% | now | 150 | 0.70% |
| it's | 242 | 0.60% | work | 355 | 0.60% | home | 102 | 0.50% | it | 147 | 0.70% |
| now | 232 | 0.60% | now | 331 | 0.60% | it's | 99 | 0.50% | home | 127 | 0.60% |
| you | 219 | 0.50% | i'm | 323 | 0.60% | me | 97 | 0.50% | some | 120 | 0.50% |
| this | 218 | 0.50% | like | 311 | 0.60% | this | 94 | 0.50% | i'm | 116 | 0.50% |
| home | 207 | 0.50% | this | 310 | 0.60% | you | 94 | 0.50% | me | 114 | 0.50% |
| me | 206 | 0.50% | me | 292 | 0.50% | now | 93 | 0.50% | it's | 108 | 0.50% |
| i'm | 194 | 0.50% | back | 285 | 0.50% | have | 88 | 0.50% | you | 104 | 0.50% |

**Appendix 5.**

**Figure 9a.** Tag clouds of stress tweets in New York.

**Figure 9b.** Tag clouds of relaxation tweets in New York.

**Figure 10a.** Tag clouds of stress tweets in Los Angeles.

**Figure 10b.** Tag clouds of relaxation tweets in Los Angeles.

**Figure 11a.** Tag clouds of stress tweets in San Diego.

**Figure 11b.** Tag clouds of relaxation tweets in San Diego.

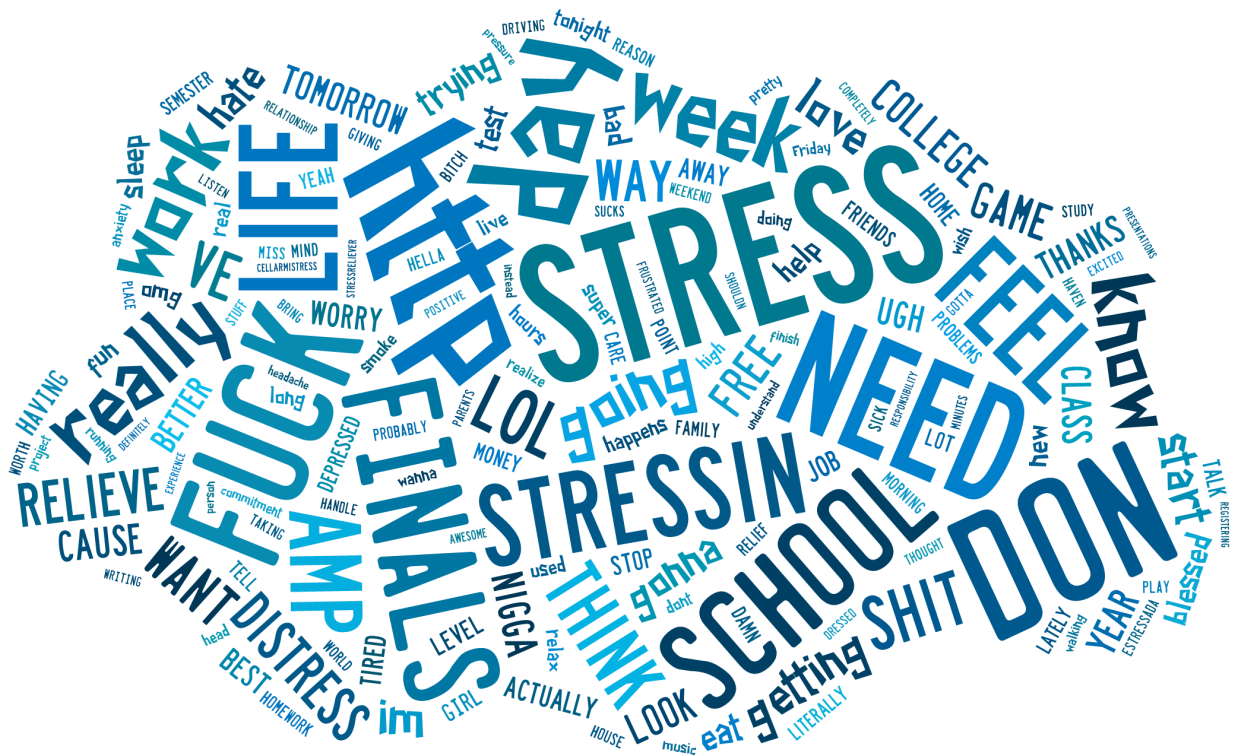

**Figure 12a.** Tag clouds of stress tweets in San Francisco.

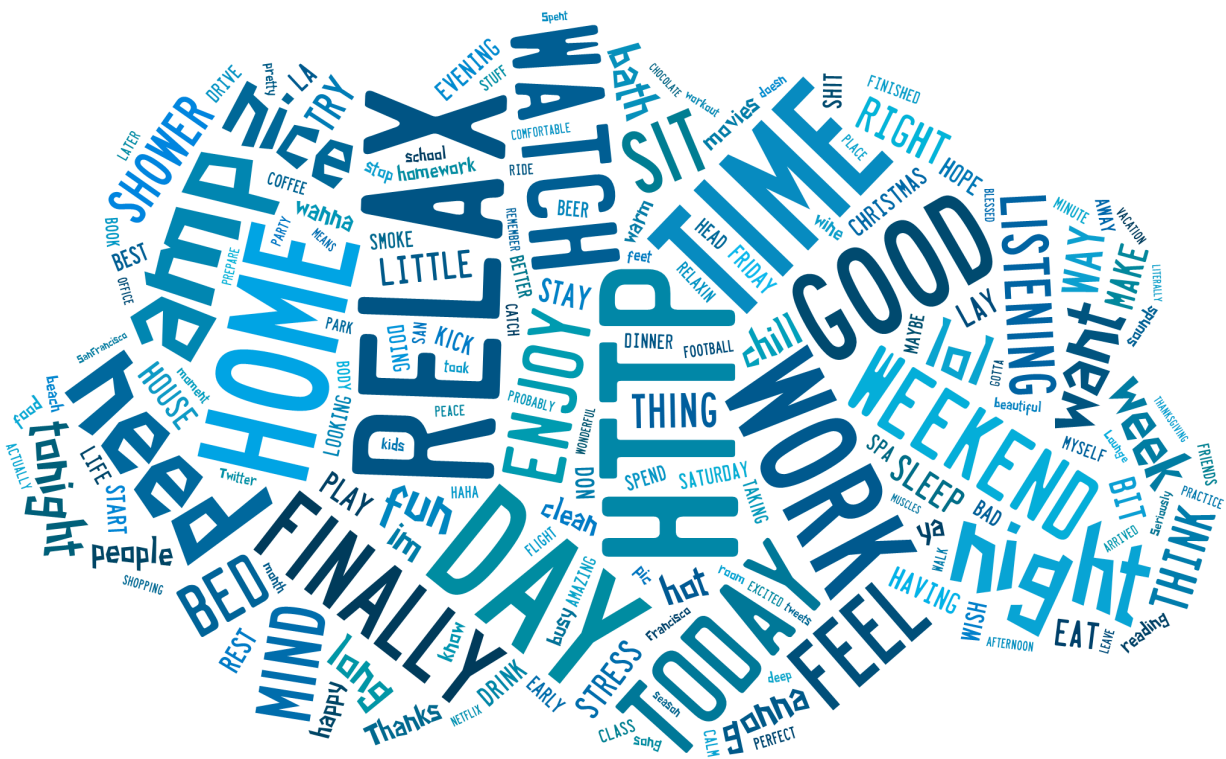

**Figure 12b.** Tag clouds of relaxation tweets in San Francisco.